\documentclass{article}

\usepackage{microtype}
\usepackage{graphicx}
\usepackage{subcaption}
\usepackage{booktabs} 
\usepackage{multirow}

\usepackage{hyperref}

\usepackage[accepted]{icml2026}

\usepackage{amsmath}
\usepackage{amssymb}
\usepackage{amsfonts}
\usepackage{mathtools}
\usepackage{amsthm}
\usepackage{algorithm}
\usepackage{algorithmic}
\usepackage{nicefrac}
\usepackage{xcolor}

\usepackage[capitalize,noabbrev]{cleveref}

\theoremstyle{plain}

\theoremstyle{definition}

\theoremstyle{remark}

\graphicspath{{./assets/}}

\definecolor{darkblue}{rgb}{0, 0, 0.5}
\hypersetup{colorlinks=true, citecolor=darkblue, linkcolor=darkblue, urlcolor=darkblue}

\newcommand{\eg}{\textit{e.g.}}
\newcommand{\ie}{\textit{i.e.}}

\icmltitlerunning{Jacobian-guided Noise Injection for Quantization Robustness}

\begin{document}

\twocolumn[
  \icmltitle{Jacobian-guided Noise Injection for Quantization Robustness \\
    in Large Language Models}


  \icmlsetsymbol{equal}{*}

  \begin{icmlauthorlist}
    \icmlauthor{Deepanshu Pandey}{Amazon}
    \icmlauthor{Arnav Chavan}{Amazon}
    \icmlauthor{Nahush Lele}{Amazon}
    \icmlauthor{Sankalp Dayal}{Amazon}
    \icmlauthor{Deepak Gupta}{Amazon}
  \end{icmlauthorlist}

  \icmlkeywords{Quantization, Large Language Models, Softmax, Jacobian Regularization, Noise Injection}

  \vskip 0.3in
]

\icmlaffiliation{Amazon}{Amazon}
\icmlcorrespondingauthor{Deepanshu Pandey}{deepnsp@amazon.com}
\printAffiliationsAndNotice{}

\begin{abstract}
Quantization of Large Language Models (LLMs) is often hindered by the sensitivity of the self-attention mechanism to discretization errors. We identify the softmax operator as a bottleneck for quantization stability due to its sensitivity to outliers and state-dependent Jacobian. We theoretically establish that suppressing the norm of this Jacobian helps in bounding quantization-induced performance degradation. Based on this, we propose Jacobian-Guided Noise Injection, a training strategy that injects zero-mean Gaussian noise into pre-attention logits, with variance derived directly from the Jacobian Frobenius norm. Unlike prior approaches that rely on heuristic or penalise jacobian directly, our method provides a way to identify the optimal noise variance based on the local attention sensitivity. We evaluate the method on SOTA LLM architectures, where it demonstrates improved robustness over popular PTQ methods. Empirical analysis reveals that the proposed method gives up to +37\% relative gains on Top-1 accuracy on ImageNet-1K for SigLIP and improves relative perplexity by upto 40\% on WikiText for language models in low bit quantisation settings, proving the efficacy of the approach.

\end{abstract}

\section{Introduction}
\label{sec:introduction}

Large Language Models (LLMs) have achieved remarkable success across a wide range of natural language tasks~\cite{touvron2023llama,touvron2023llama2,qwen2024qwen25}. However, deploying these models efficiently remains challenging due to their substantial computational and memory requirements~\cite{gholami2022survey,tang2024survey}. Quantization offers a promising path to efficient deployment by reducing the precision of weights and activations~\cite{jacob2018quantization}. For instance, 4-bit weight quantization can reduce memory footprint by 4$\times$ compared to FP16 models and achieve over 3$\times$ inference speedup, even enabling deployment of 70B parameter models on mobile GPUs~\cite{lin2024awq}.

However, naive quantization often leads to significant performance degradation~\cite{frantar2023gptq,lin2024awq}, due to approximation noise and rounding errors in lower-precision that perturb intermediate computations~\cite{micikevicius2018mixed}. The degradation becomes more severe at lower bit-widths (e.g., 4-bit or below) due to reduced representational capacity and the presence of activation outliers . Recent studies show that quantization-induced errors disproportionately affect mathematical reasoning, multi-step planning, and long-context tasks, motivating the development of robust quantization-aware and activation-aware methods \citep{li2025quantizationmeetsreasoningexploring,lin2026awqactivationawareweightquantization,xiao2024smoothquantaccurateefficientposttraining} 
\begin{figure}[t]
\centering
\includegraphics[width=\linewidth]{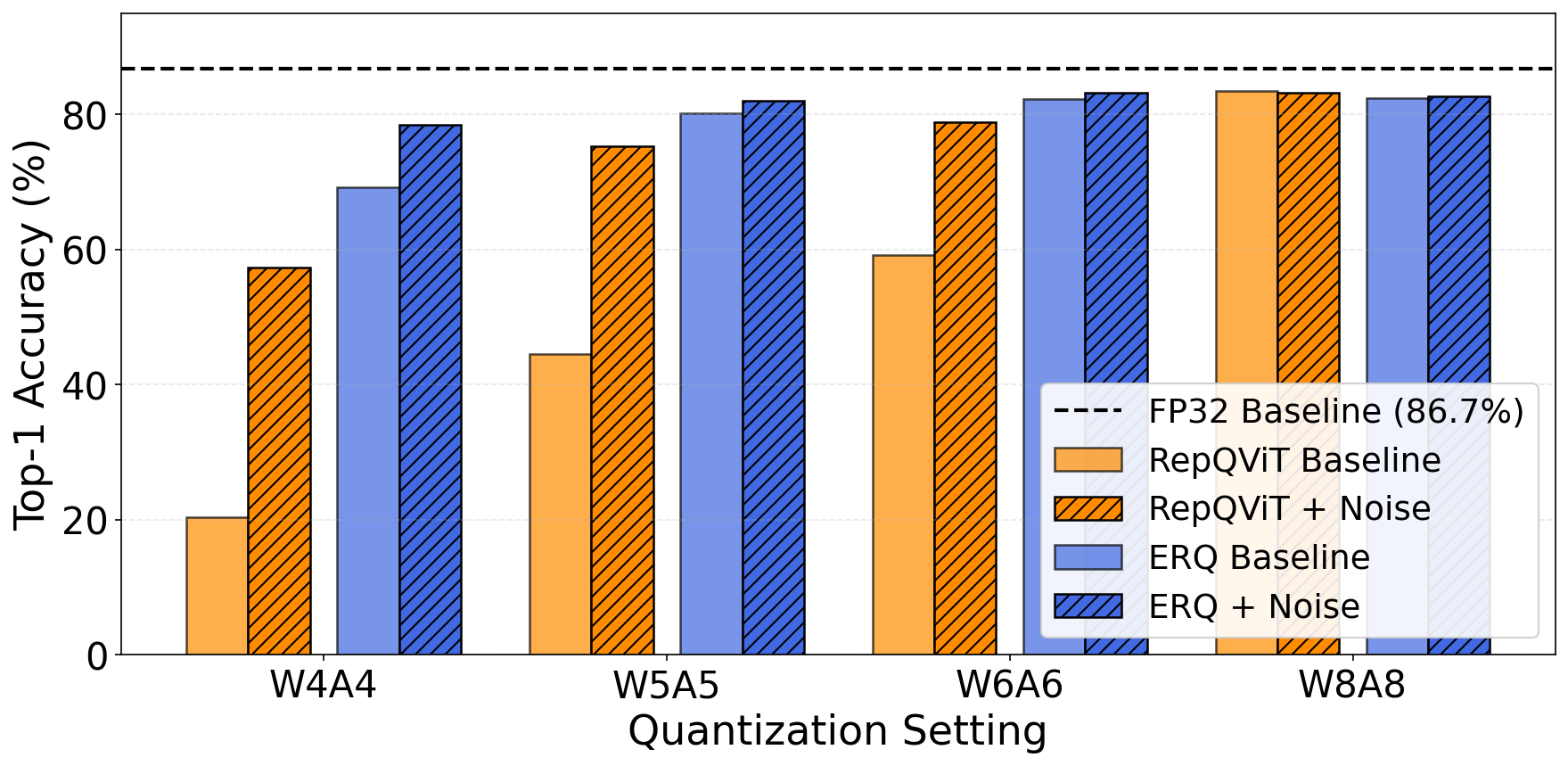}
\caption{PTQ results on ImageNet-1K Top-1 accuracy for SigLIP base 16-384 with ERQ and RepQViT.}
\label{fig:siglip_results}
\end{figure}
A critical observation is that the self-attention mechanism in Transformers~\cite{vaswani2017attention} is particularly sensitive to quantization errors, due to the presence of highly sensitive operations like softmax, normalization etc. We observed this phenomenon in quantised model deployment (Figure \ref{fig:quant_cosine_similarity_attn}) where the error propagation was highly pronounced for the attention layers, leading to major diversion from expected activation values. Unlike linear layers where error propagation is bounded by fixed weight matrices, the Softmax operator in attention exhibits \textit{state-dependent} sensitivity that varies dramatically based on the input distribution~\cite{kim2021ibert,lin2022fqvit}. The problem is especially severe when logits contain outliers or large magnitudes, since quantization errors in these values are amplified \textit{exponentially} through the softmax gate due to its exponential nonlinearity~\cite{xiao2023smoothquant,dettmers2022llmint8}. A small perturbation to a large logit produces a disproportionately large change in the output probability, causing catastrophic error propagation. This creates unpredictable error amplification that standard quantization techniques fail to address.

In this work, we analyse the quantization error propagation through the softmax operator and derive conditions for bounding this error using norm of the softmax jacobian. Furthermore, we propose Jacobian-guided noise injection, a training strategy to improve the downstream performance of quantised model (Figure \ref{fig:siglip_results}). Our key contributions are:
\begin{enumerate}
    \item We analyse the relationship between  spectral norm of the softmax jacobian and quantization error amplification, and show that minimizing expected loss under logit perturbation implicitly regularizes this norm.
    \item We derive an expression to approximate the Jacobian Frobenius norm and use it to calibrate noise injection variance, providing a simpler alternative to heuristic approaches.
    \item We demonstrate that our Jacobian-guided noise injection method improves quantization robustness on multiple LLM architectures and quantisation settings, recovering up to 37\% accuracy in a W4A4 with zero inference overhead.
\end{enumerate}

\section{Methodology}
\label{sec:methodology}

In this section, we identify the self-attention Softmax operator as the bottleneck for quantization stability in Transformers~\cite{vaswani2017attention}. We theoretically analyse the conditions required to bound this quantization error and observe that these conditions can be satisfied via an implicit Hessian regularization~\cite{bishop1995training}. Finally, to apply this regularisation in practical scenarios, we propose a fine-tuning framework to achieve robust quantization.

\subsection{Sensitivity of Softmax Jacobian}
\label{subsec:bottleneck}

In a standard Transformer, the self-attention mechanism computes attention probabilities from the pre-activation logits. To analyze error propagation, let $z \in \mathbb{R}^N$ denote a single row vector of the pre-activation logit matrix for a given query (\ie, $z_i = q^T k_i / \sqrt{d}$). The corresponding attention probability vector $a \in \mathbb{R}^N$ is computed via the Softmax function:
\begin{equation} \label{eq:softmax}
    a = S(z)  = \frac{e^{z_i}}{\sum_{j=1}^{N} e^{z_j}}
\end{equation}
When the model is deployed in a quantized format, the discretization of weights and activations introduces a bounded perturbation $\delta \in \mathbb{R}^N$ into the logits, such that the quantized pre-activations are $z_q = z + \delta$. The error propagated into the attention distribution is:
\begin{equation} \label{eq:delta_a}
    \Delta a = S(z + \delta) - S(z)
\end{equation}

Using first-order Taylor expansion, we approximate this error via the jacobian $J_S(z) \in \mathbb{R}^{N \times N}$:
\begin{equation} \label{eq:jacobian_bound}
    \Delta a \approx J_S(z) \delta \implies ||\Delta a||_2 \le ||J_S(z)||_2 ||\delta||_2
\end{equation}

The critical vulnerability lies in the formulation of the softmax jacobian $J_S(z)$:
\begin{equation} \label{eq:jacobian_def}
    J_S(z)_{i,j} = \partial a_i / \partial z_j = a_i (\mathbf{1}_{i=j} - a_j)
\end{equation}

Unlike linear layers where input Jacobian is a constant weight matrix (\eg, $\nabla_X (XW) = W^T$), $J_S(z)$ is dense and strictly state-dependent. The structure of the Jacobian provides insight into the sensitivity of the softmax output with respect to its input logits. When attention is concentrated on a single token, it results in safe, saturated regions where the Jacobian norm approaches zero, compared to highly sensitive regions where equal mass is over 2 or more tokens and quantization errors are aggressively amplified (norm decreases as mass is distributed across more tokens). Standard fine-tuning objectives do not take this into account. Consequently, an unregularized model may learn pre-activations that rest in these sensitive regions, maximizing the $||J_S(z)||_2$. In these regimes, even a minimal quantization error $||\delta||_2$ triggers an unpredictable, exponential amplification of $\Delta a$, leading to catastrophic task degradation. 
The proposed noise injection strategy aims to implicitly induce this regularisation during training by making the injected noise a function of the jacobian state.

\begin{figure}[t]
\centering
    \includegraphics[trim={32.5cm 1cm 32cm 27cm}, clip, width=0.8\linewidth]{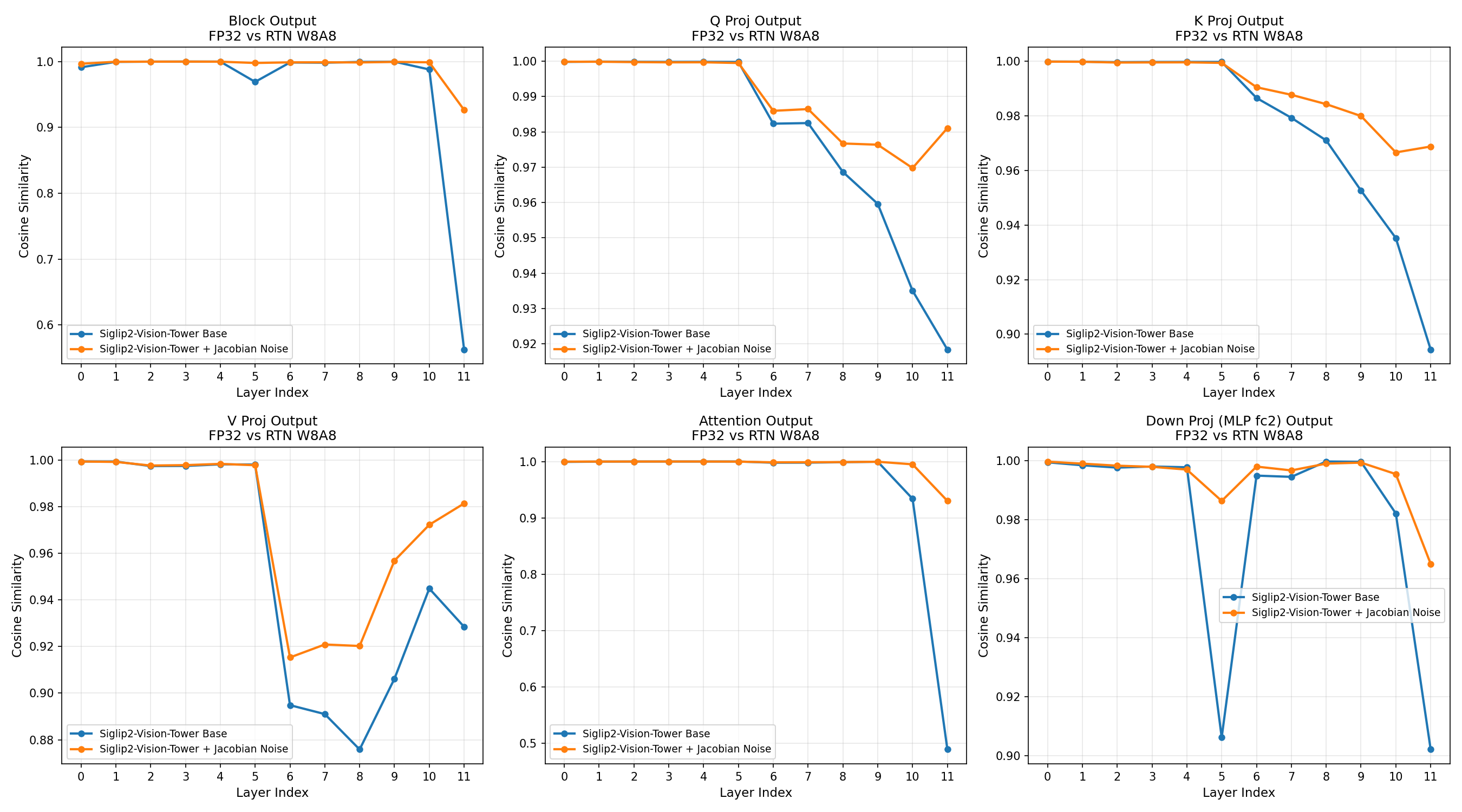}
\caption{Activation cosine similarity for attention outputs in Siglip-base-384.}
\label{fig:quant_cosine_similarity_attn}
\end{figure}

\subsection{Constraining the Softmax Jacobian}
\label{subsec:theoretical_req}

To strictly bound the quantization error $||\Delta a||_2$, we must constrain the spectral norm of the Jacobian, $||J_S(z)||_2$. Let $\mathcal{L}(z) = (l \circ S)(z)$ represent the end-to-end loss as a function of the pre-activation logits.
Directly penalizing $\|J_S(z)\|_2$ is computationally prohibitive as it is not a static parameter but a state-dependent matrix. We can establish a theoretical bound on the Jacobian by regularizing the logit Hessian, $\nabla^2_z \mathcal{L}(z)$. Applying the multivariate chain rule, the logit Hessian decomposes as:
\begin{equation} \label{eq:hessian_full}
    \nabla^2_z \mathcal{L}(z) = J_S(z)^T \nabla^2_a \ell(a) J_S(z) + \sum_{k=1}^N \frac{\partial \ell}{\partial a_k} \nabla^2_z S_k(z)
\end{equation}
Let $H_a = \nabla^2_a \ell(a)$ denote the activation Hessian. Using Gauss-Newton approximation and ignoring the second-order residual term, the relationship simplifies to:
\begin{equation} \label{eq:gauss_newton}
    \nabla^2_z \mathcal{L}(z) \approx J_S(z)^T H_a J_S(z)
\end{equation}
Near a local optimum, the loss landscape is locally convex, meaning $H_a$ is positive semi-definite ($H_a \succeq 0$). Let $\lambda_{\min} > 0$ denote the smallest positive eigenvalue of $H_a$. By the properties of positive semi-definite matrices, we can bound the trace of the Hessian:
\begin{equation} \label{eq:trace_bound} \text{Tr}\left(J_S(z)^T H_a J_S(z)\right) \ge \lambda_{\min} \text{Tr}\left(J_S(z)^T J_S(z)\right)
\end{equation}

\begin{algorithm}[t]
\caption{Jacobian-Guided Noise Injection}
\label{alg:jacobian_noise}
\begin{algorithmic}[1]
\REQUIRE Model $\mathcal{M}$, update interval $T$, scale factor $\alpha$
\FOR{each training step $t$}
    \IF{$t \mod T = 0$}
        \STATE Forward pass to compute attention $P^{(\ell)}$ per layer
        \FOR{each layer $\ell$}
            \STATE Compute $||J_S^{(\ell)}||_F^2$ using Eq.~\ref{eq:jacobian_closed_form}
            \STATE Update $\sigma_i^{(\ell)} \leftarrow \alpha \cdot \sqrt{\mathbb{E}_{b,h}[||J_S||_{F,i}^2]}$
            \STATE Clamp: $\sigma \leftarrow \text{clamp}(\sigma, \sigma_{\min}, \sigma_{\max})$
        \ENDFOR
    \ENDIF
    \STATE Compute logits: $Z = QK^T / \sqrt{d}$
    \STATE Sample noise: $\epsilon_i \sim \mathcal{N}(0, \sigma_i^2)$
    \STATE Perturb: $\tilde{Z} = Z + \epsilon$
    \STATE Apply softmax: $\tilde{A} = \text{Softmax}(\tilde{Z})$
    \STATE Compute loss $\mathcal{L}(\tilde{A})$ and backpropagate
\ENDFOR
\end{algorithmic}
\end{algorithm}

By the definition of the Frobenius norm, $\text{Tr}(J_S(z)^T J_S(z)) = ||J_S(z)||_F^2$. Since the spectral norm satisfies $||J_S(z)||_2^2 \le ||J_S(z)||_F^2$, we obtain:
\begin{equation} \label{eq:theoretical_condition}
    ||J_S(z)||_2 \le \sqrt{\frac{\text{Tr}(\nabla^2_z \mathcal{L}(z))}{\lambda_{\min}}}
\end{equation}
\cref{eq:theoretical_condition} directly bounds the error amplification defined in \cref{eq:jacobian_bound}. Therefore, our ideal regularization objective should penalize the trace of the logit Hessian:
\begin{equation} \label{eq:ideal_obj_formal}
    \mathcal{L}_{\text{ideal}}(z) = \mathcal{L}(z) + \lambda \cdot \text{Tr}(\nabla^2_z \mathcal{L}(z))
\end{equation}
where $\lambda > 0$ controls regularization strength. However, computing this requires continuous backward passes of second-order derivatives, which is computationally intractable for large Transformers. \cref{subsec:stochastic_perturbation} shows how to efficiently approximate this penalty using only first-order gradients.

\subsection{Stochastic Perturbation}

We now derive a first-order approximation to $\mathcal{L}_{\text{ideal}}$. Consider modifying the objective to minimize the expected loss under a continuous, zero-mean perturbation $\epsilon \in \mathbb{R}^N$ applied directly to the logits: $  \mathbb{E}_{\epsilon} [\mathcal{L}(z + \epsilon)]$ . We analyze the behavior of this objective via a second-order Taylor series expansion of the perturbed loss around the unperturbed logits $z$:
\begin{equation} \label{eq:taylor_expansion}
    \mathcal{L}(z + \epsilon) = \mathcal{L}(z) + \nabla_z \mathcal{L}(z)^T \epsilon + \frac{1}{2} \epsilon^T \nabla^2_z \mathcal{L}(z) \epsilon + \mathcal{O}(||\epsilon||^3)
\end{equation}
We explicitly define the perturbation $\epsilon$ as isotropic Gaussian noise sampled from $\mathcal{N}(0, \sigma^2 I)$. By leveraging its statistical properties ($\mathbb{E}[\epsilon] = 0$ and $\mathbb{E}[\epsilon \epsilon^T] = \sigma^2 I$), taking the expectation of the Taylor expansion causes the first-order gradient term to vanish completely:
\begin{equation} \label{eq:expected_loss}
    \mathbb{E}_{\epsilon \sim \mathcal{N}(0, \sigma^2 I)} [\mathcal{L}(z + \epsilon)] \approx \mathcal{L}(z) + \frac{\sigma^2}{2} \text{Tr}(\nabla^2_z \mathcal{L}(z))
\end{equation}
This derivation yields a critical similarity: minimizing the expected loss under Gaussian logit perturbation matches our requirement derived in \cref{subsec:theoretical_req}. Thus, the perturbation variance $\sigma^2$ acts as the implicit regularization strength $\lambda$ in \cref{eq:ideal_obj_formal}, with the correspondence $\lambda = \sigma^2 / 2$.

\subsection{Jacobian-Guided Noise Injection}
\label{subsec:proposed_method}

\begin{table*}[t]
\centering
\small
\caption{Zero shot PTQ results for Llama-3.2-3B and Qwen2.5-3B. Acc is average accuracy of over 7 benchmarks ($\uparrow$).}
\label{tab:awq_results}
\begin{tabular}{lllrrrrrr}
\toprule
& & & \multicolumn{2}{c}{\textbf{AWQ}} & \multicolumn{2}{c}{\textbf{GPTQ}} & \multicolumn{2}{c}{\textbf{SpinQuant}} \\
\textbf{Model} & \textbf{Precision} & \textbf{Variant} & \textbf{PPL} $\downarrow$ & \textbf{Acc} $\uparrow$ & \textbf{PPL
} $\downarrow$ & \textbf{Acc} $\uparrow$ & \textbf{PPL} $\downarrow$ & \textbf{Acc} $\uparrow$ \\
\midrule

\multirow{8}{*}{Llama-3.2-3B}
  & \multirow{2}{*}{W4A4}
    & Base    & 105.06 & 42.06 & \textbf{454.35} & 40.31 & \textbf{10.72} & \textbf{60.25} \\
  & & Ours  & \textbf{104.34} & \textbf{42.20} & 496.38 & \textbf{41.00} & 11.18 & 60.15 \\
\cmidrule(l){2-9}
  & \multirow{2}{*}{W4A8}
    & Base    &  9.96 & 66.65 & 37.40 & 64.81 & \textbf{8.31} & 64.48 \\
  & & Ours  & \textbf{9.86} & \textbf{67.35} &  \textbf{31.47} & \textbf{66.36} &  8.47 & \textbf{65.17} \\
\cmidrule(l){2-9}
  & \multirow{2}{*}{W6A6}
    & Base    & 10.80 & 64.07 & \textbf{10.87} & 63.69 & \textbf{8.03} & 65.98 \\
  & & Ours  & \textbf{10.55} & \textbf{64.85} &  11.12 & \textbf{64.16} &  8.17 & \textbf{66.81} \\
\cmidrule(l){2-9}
  & \multirow{2}{*}{W8A8}
    & Base    &  9.73 & 66.62 &  \textbf{9.69} & 66.69 & \textbf{7.97} & 65.95 \\
  & & Ours  & \textbf{9.68} & \textbf{67.55} &   9.86 & \textbf{67.25} &  8.12 & \textbf{66.72} \\
\midrule

\multirow{8}{*}{Qwen2.5-3B}
  & \multirow{2}{*}{W4A4}
    & Base    &    8121.24 & 40.58 & 5172.39 & 38.58 & 10.49 & 60.34 \\
  & & Ours  & \textbf{3281.04} & \textbf{41.56} & \textbf{4207.85} & \textbf{38.96} & \textbf{10.33} & \textbf{62.62} \\
\cmidrule(l){2-9}
  & \multirow{2}{*}{W4A8}
    & Base    & \textbf{10.80} & 67.84 & 14.13 & 66.65 &  8.55 & \textbf{67.35} \\
  & & Ours  &  11.01 & \textbf{68.38} &    \textbf{12.21} & \textbf{66.11} & \textbf{8.35} & 65.66  \\
\cmidrule(l){2-9}
  & \multirow{2}{*}{W6A6}
    & Base    &  15.29 & 65.12 &  \textbf{28.42} & 59.97 &  8.36 & 67.25 \\
  & & Ours  & \textbf{13.63} & \textbf{65.38} &  21.52 & \textbf{60.10} & \textbf{8.18} & \textbf{67.44} \\
\cmidrule(l){2-9}
  & \multirow{2}{*}{W8A8}
    & Base    & 10.80 & 67.84 & 10.98 & 67.55 &  8.32 & \textbf{67.66} \\
  & & Ours  &  10.80 & \textbf{68.38} &  \textbf{10.82} & \textbf{67.61} & \textbf{8.11} & 67.45  \\
\bottomrule
\end{tabular}
\end{table*}

\label{subsec:stochastic_perturbation}

\cref{subsec:stochastic_perturbation} established that Gaussian noise injection with variance $\sigma^2$ implicitly regularizes the Hessian trace with strength $\lambda = \sigma^2/2$. However, a fixed global $\sigma$ treats all layers and positions uniformly, ignoring the fact that the Jacobian norm (and hence quantization sensitivity) varies dramatically across the network. Positions on sensitive regions (\cref{subsec:bottleneck}) require stronger regularization than those in saturated regions. We therefore propose deriving the noise variance directly from the local Jacobian Frobenius norm, yielding an adaptive scheme that concentrates regularization precisely where it is needed. For softmax output $p = S(z)$, the Jacobian $J_S = \text{diag}(p) - pp^T$ admits a closed-form Frobenius norm:
\begin{equation} \label{eq:jacobian_closed_form}
    ||J_S||_F^2 = ||p||_2^2 - 2||p||_3^3 + ||p||_2^4
\end{equation}
where $||p||_k = (\sum_i p_i^k)^{1/k}$. This expression requires only element-wise operations and reductions (no explicit Jacobian materialization) adding negligible overhead to the forward pass. We set the noise standard deviation proportional to the Jacobian norm: $\sigma_i = \alpha \cdot \sqrt{||J_S||_{F,i}^2}$, where $\alpha$ is a scaling hyperparameter. This ensures that noise \emph{variance} scales linearly with the Jacobian norm squared, preserving the correspondence $\lambda \propto \sigma^2$ from \cref{eq:expected_loss}. Each query position receives noise calibrated to its local sensitivity:
\begin{equation}
    \epsilon_i \sim \mathcal{N}(0, \sigma_i^2), \quad \sigma_i = \alpha \cdot \sqrt{\mathbb{E}_{b,h}[||J_S||_{F,i}^2]}
\end{equation}
We adopt rowwise (per-position) noise rather than a static global value because the Jacobian norm varies substantially even within a single attention head. Rowwise injection applies stronger perturbation to high-sensitivity positions while leaving saturated positions largely undisturbed, directly targeting the transitional ridges identified in \cref{subsec:bottleneck}. \cref{alg:jacobian_noise} summarizes the training procedure. Noise parameters are updated periodically (every $T$ steps) to track evolving attention patterns, and clamped to $[\sigma_{\min}, \sigma_{\max}]$ for numerical stability.
At inference time, noise injection is disabled ($\sigma = 0$). The model retains the flattened local geometry learned during training (analogous to the effect of dropout~\cite{srivastava2014dropout}) where attention distributions have been pushed away toward robust, saturated regions, ensuring quantization robustness.

\section{Experiments}

\subsection{Experimental Setup}
\label{sec:experiments}

We evaluate the effectiveness of Post Training Quantisation (PTQ) with Jacobian-guided noise injection on state-of-the-art language and vision-language models. We experiment with two open-source language models: Llama-3.2-3B~\cite{grattafiori2024llama3} and Qwen2.5-3B~\cite{qwen2024qwen25} fine-tuned on the Alpaca dataset~\cite{taori2023alpaca}. For vision-language, we use SigLIP base 16-384~\cite{zhai2023siglip} trained on ImageNet-1K~\cite{deng2009imagenet} classification task. We compare the baseline (standard fine-tuning without noise) against our Jacobian-guided noise injection under several low-bit quantisation settings. For PTQ, we employ AWQ~\cite{lin2024awq},  GPTQ~\cite{frantar2023gptq} and SpinQuant for language models, and ERQ and RepQViT~\cite{li2023repqvit} for vision task. Models are assessed in a zero shot setting on MMLU~\cite{hendrycks2021mmlu}, HellaSwag~\cite{zellers2019hellaswag}, Winogrande~\cite{sakaguchi2020winogrande}, PIQA~\cite{bisk2020piqa}, TruthfulQA~\cite{lin2022truthfulqa}, and WikiText~\cite{merity2017pointer} perplexity. See \cref{app:experimental_details} for details.

\subsection{Main Results}


\cref{tab:awq_results} present results for Llama and Qwen models under PTQ setting using AWQ, GPTQ and Spinquant. Jacobian-guided noise injection yields consistent improvements across quantization levels for both Llama-3.2-3B and Qwen2.5-3B. Notably, the method improves quantized performance without degrading full-precision accuracy significantly (see \cref{app:fp16_noise}), indicating that noise injection learns representations that are inherently more robust to discretization error.

\begin{figure*}[t]
\centering
\begin{subfigure}[b]{0.45\linewidth}
    \includegraphics[width=\linewidth]{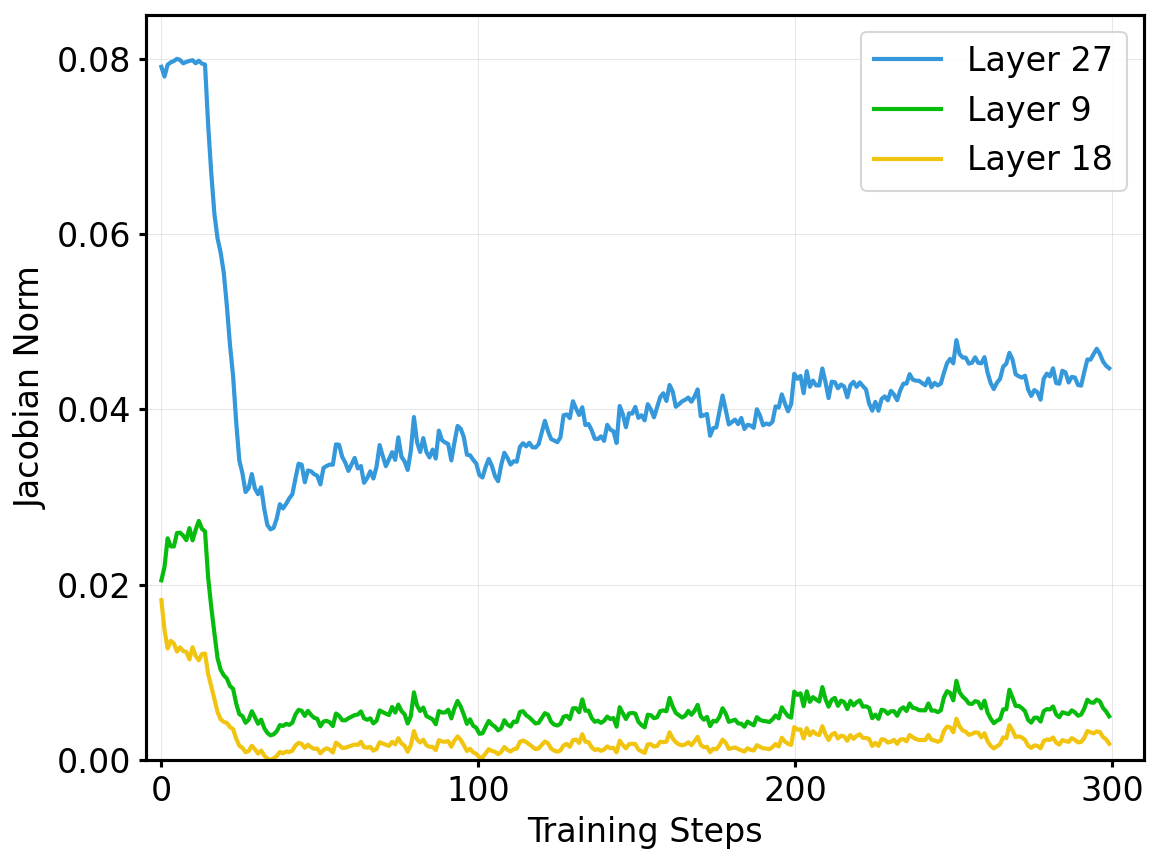}
    \caption{}
    \label{fig:jacobian_norm_training}
\end{subfigure}
\hfill
\begin{subfigure}[b]{0.43\linewidth}
    \includegraphics[width=\linewidth]{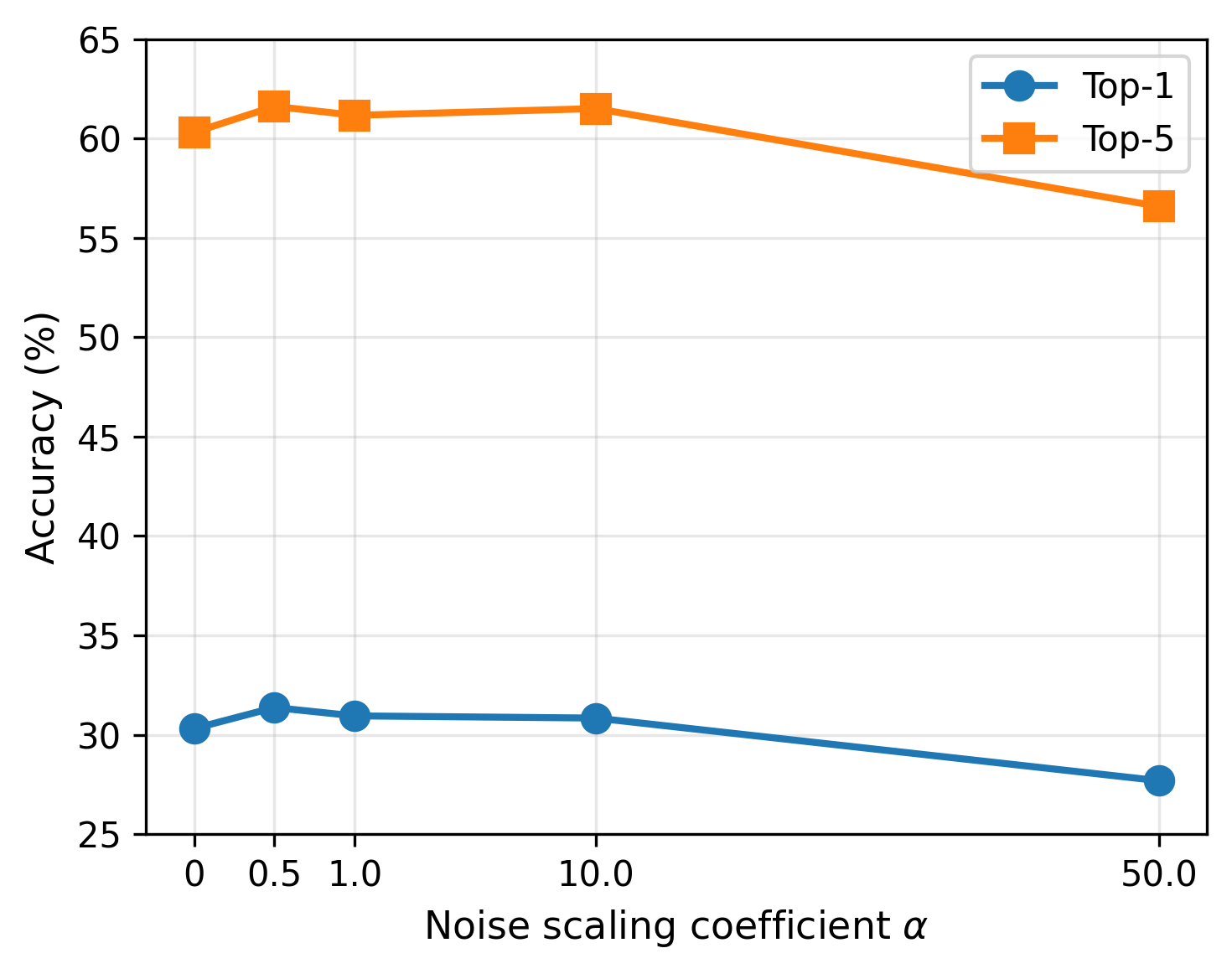}
    \caption{}
    \label{fig:alpha_sensitivity}
\end{subfigure}
\caption{(a) Jacobian norm during training for Qwen2.5-3B layers 9, 18, and 27. (b) Effect of scaling coefficient $\alpha$ on SigLIP W4A4 QAT (ImageNet-1K zero-shot accuracy). Moderate scaling ($\alpha \in [0.5, 1.0]$) yields better results compared to no noise ($\alpha = 0 $).}
\label{fig:jacobian_norm}
\end{figure*}

\subsection{Analysis}

\label{sec:analysis}
We structure our analysis around the following research questions to systematically evaluate the effectiveness and generalizability of Jacobian-guided noise injection for post-training quantisation: 

\textbf{RQ1.} \textit{Are performance gains generalizable across model families and quantization strategies?}

The consistent gains across three model families (Llama, Qwen, SigLIP~\cite{zhai2023siglip}) and 4 PTQ methods (AWQ~\cite{lin2024awq}, GPTQ~\cite{frantar2023gptqaccurateposttrainingquantization}, ERQ~\cite{zhong2025accurateposttrainingquantizationvision}, RepQViT~\cite{li2023repqvit}) demonstrate that Jacobian-guided noise injection learns quantization-robust representations rather than overfitting to a specific architecture or quantization scheme. \cref{fig:siglip_results} evaluates generalization to vision transformers~\cite{dosovitskiy2021image} using SigLIP base 16-384 on ImageNet-1K~\cite{deng2009imagenet} classification. With ERQ, noise injection recovers up to +9\% accuracy at W4A4; with RepQViT, gains reach +37\% at the same bit-width. The improvements scale inversely with bit-width, highlighting the fact that aggressive quantization benefits most from noise-based regularization. Beyond PTQ, we also evaluate SigLip model under quantization-aware training (QAT). \cref{fig:alpha_sensitivity} shows that noise injection ($\alpha > 0 $) improves over QAT baseline performance ($\alpha = 0 $) for SigLIP, confirming that the method can be used with existing QAT pipelines~\cite{choi2018pact} and is not restricted to PTQ settings.


\textbf{RQ2.} \textit{ How does noise injection affect attention maps and logit distributions?}

\begin{figure*}[t]
\centering
    \includegraphics[width=0.72\linewidth]{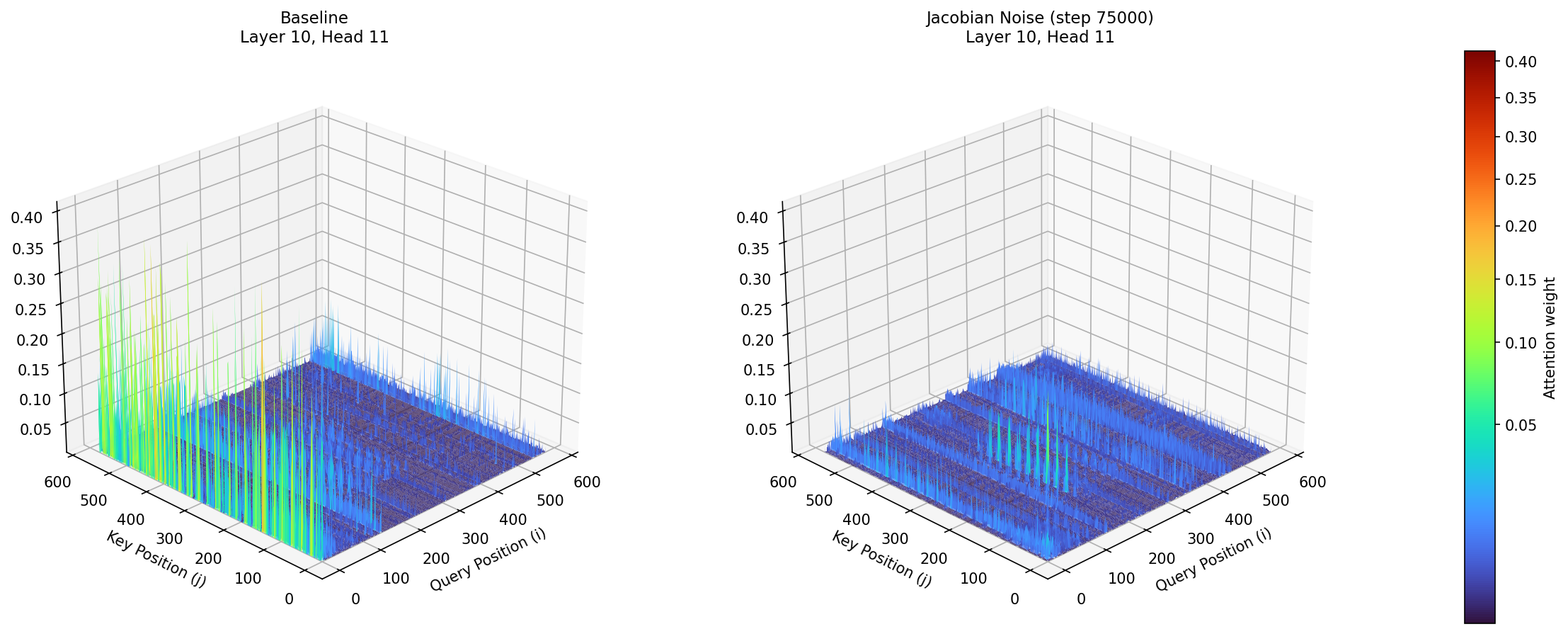}
    \label{fig:attention_maps_base}
\caption{Attention maps for SigLIP (layer 10). Noise-trained models exhibit more diffused attention patterns across key positions.}
\label{fig:attention_maps}
\end{figure*}

\cref{fig:jacobian_norm} shows the Jacobian norm decreasing and stabilizing during noise-based training, providing empirical evidence for noise-based Jacobian desensitisation. \cref{fig:attention_maps} compares attention maps for SigLIP-base-384 trained with and without noise. Noise-trained models exhibit more diffuse attention patterns achievable only in smaller Jacobian norm regions, consistent with the theoretical predictions in \cref{sec:methodology}. We observe this phenomenon of diffused attention across all model layers, pointing towards the stochasticity-based robustness and outlier reduction introduced due to injected noise~\cite{hendrycks2019benchmarking}. It should be noted that, while zero Jacobian norm is achievable when probability mass concentrates on a single token, this regime is somewhat problematic since: (1) sparse attention degenerates to near-constant outputs for all kinds of inputs, and (2) large logit magnitudes amplify quantization errors exponentially through the softmax gate~\cite{guo2017calibration}. Noise injection prevents this collapse and the stochastic perturbation discourages extreme weight concentration on any single token, analogous to the effect of dropout~\cite{srivastava2014dropout} on model training. The resulting distributed attention provides innate robustness to quantization error propagation.

\textbf{RQ3.} \textit{ How does the noise scaling coefficient $\alpha$ affect quantized model performance?}

Figure \ref{fig:alpha_sensitivity} presents ImageNet-1K zero-shot accuracy for SigLIP at W4A4 QAT setting across different values of $\alpha$. We observer that moderate scaling ($\alpha \in [0.5, 1.0]$) yields the best results, with $\alpha=0.5$ achieving 31.37\% Top-1 accuracy compared to 30.31\% for the baseline (+1.06\%). Performance degrades rapidly after this and the model performance is the lowest at $\alpha=50.0$ (27.69\%), indicating that excessive noise disrupts learning due to logits being pushed beyond representable ranges. We observed similar trend for both PTQ and QAT settings, where performance improves and then rapidly degrading with increasing noise.

\section{Related Work}
\label{sec:related}

Quantisation has emerged as a dominant approach for deploying large models on resource-constrained environments~\cite{gholami2022survey,tang2024survey,dettmers2022llmint8}. Post-training quantization (PTQ) methods~\cite{frantar2023gptq,lin2024awq,xiao2023smoothquant,dettmers2022llmint8} like GPTQ and AWQ compress models to 4-8 bits using second-order information, salient weight protection, or outlier handling. Quantization-aware training (QAT) and mixed-precision approaches~\cite{dong2019hawq,wang2019haq} take an alternative route by incorporating quantization into the training loop~\cite{esser2020learned,choi2018pact}. However, both PTQ and QAT methods typically treat all layers uniformly without addressing component-specific sensitivities. The attention mechanism~\cite{vaswani2017attention} poses unique challenges for quantization due to the softmax nonlinearity, which creates error amplification that fixed quantization schemes cannot anticipate. Several works~\cite{liu2021posttraining} like I-BERT~\cite{kim2021ibert}, FQ-ViT~\cite{lin2022fqvit}, and RepQ-ViT~\cite{li2023repqvit} address this by modifications at inference time. However, these methods engineer around softmax sensitivity rather than addressing the root cause.

The connection between Jacobian norms and model robustness provides a theoretical foundation for addressing this sensitivity~\cite{goodfellow2015explaining,madry2018towards}. Contractive autoencoders~\cite{rifai2011contractive} penalize the Frobenius norm of the encoder Jacobian to learn locally invariant features.\cite{jakubovitz2018jacobian} extends this to adversarial robustness, showing that Jacobian regularization bounds sensitivity to input perturbations. More recent work~\cite{nguyen2024discriminative} demonstrated that improving feature stability under Gaussian noise implicitly reduces curvature of the softmax loss landscape, connecting noise injection to loss geometry. Noise injection as regularisation~\cite{srivastava2014dropout,wan2013regularization,bishop1995training} has also been explored in some prior works. \cite{chen2017noisy} injects annealed noise to postpone early softmax saturation during training. However, most of these methods do not address quantisation as the target objective, and use fixed or annealed noise schedules that do not account for position-dependent sensitivity.

Our work synthesizes these threads by targeting the attention softmax specifically and deriving noise variance from the Jacobian norm. Unlike prior noise injection methods that apply uniform perturbation, the proposed approach adapts noise per-position based on local sensitivity, concentrating regularization where quantization errors are most amplified, which produces models with attention distributions that are inherently robust to precision perturbations, providing benefits that transfer across quantization methods.

\section{Conclusion}
\label{sec:conclusion}

We presented Jacobian-guided Noise Injection, a noval approach to improving quantization robustness in Large Language Models. By deriving noise variance directly from the Softmax Jacobian Frobenius norm, our method provides adaptive regularization that targets the most sensitive regions of the attention softmax. Our theoretical analysis establishes clear connections between noise injection, Hessian trace regularization, and quantization error bounds. Experiments across multiple PTQ and QAT methods across multiple model families demonstrate consistent improvements. We also provide empirical evidence for the relationship between noise injection and Jacobian sensitivity through qualitative analysis. The proposed method adds no overhead at inference time, matches baseline performances at fp16 and improves quantisation robustness, making it practical for deployment in resource-constrained environments.


\section*{Impact Statement}
This paper presents work whose goal is to develop robust quantisation approaches. Our method enables aggressive low-bit quantization with reduced quality degradation, which reduce the computational cost, memory footprint, and energy required to deploy LLMs. The proposed methods are general-purpose techniques applicable to any neural network and do not introduce novel capabilities. We do not foresee any ethical concerns beyond those that are well established in the field of machine learning efficiency.

\bibliography{references}

@misc{li2025quantizationmeetsreasoningexploring,
      title={Quantization Meets Reasoning: Exploring LLM Low-Bit Quantization Degradation for Mathematical Reasoning}, 
      author={Zhen Li and Yupeng Su and Runming Yang and Congkai Xie and Zheng Wang and Zhongwei Xie and Ngai Wong and Hongxia Yang},
      year={2025},
      eprint={2501.03035},
      archivePrefix={arXiv},
      primaryClass={cs.CL},
      url={https://arxiv.org/abs/2501.03035}, 
}

@misc{liu2025spinquantllmquantizationlearned,
      title={SpinQuant: LLM quantization with learned rotations}, 
      author={Zechun Liu and Changsheng Zhao and Igor Fedorov and Bilge Soran and Dhruv Choudhary and Raghuraman Krishnamoorthi and Vikas Chandra and Yuandong Tian and Tijmen Blankevoort},
      year={2025},
      eprint={2405.16406},
      archivePrefix={arXiv},
      primaryClass={cs.LG},
      url={https://arxiv.org/abs/2405.16406}, 
}

@misc{lin2026awqactivationawareweightquantization,
      title={AWQ: Activation-aware Weight Quantization for LLM Compression and Acceleration}, 
      author={Ji Lin and Jiaming Tang and Haotian Tang and Shang Yang and Wei-Ming Chen and Wei-Chen Wang and Guangxuan Xiao and Xingyu Dang and Chuang Gan and Song Han},
      year={2026},
      eprint={2306.00978},
      archivePrefix={arXiv},
      primaryClass={cs.CL},
      url={https://arxiv.org/abs/2306.00978}, 
}

@misc{xiao2024smoothquantaccurateefficientposttraining,
      title={SmoothQuant: Accurate and Efficient Post-Training Quantization for Large Language Models}, 
      author={Guangxuan Xiao and Ji Lin and Mickael Seznec and Hao Wu and Julien Demouth and Song Han},
      year={2024},
      eprint={2211.10438},
      archivePrefix={arXiv},
      primaryClass={cs.CL},
      url={https://arxiv.org/abs/2211.10438}, 
}

@inproceedings{rifai2011contractive,
  title={Contractive Auto-Encoders: Explicit Invariance During Feature Extraction},
  author={Rifai, Salah and Muller, Xavier and Glorot, Xavier and Mesnil, Gregoire and Bengio, Yoshua and Vincent, Pascal},
  booktitle={International Conference on Machine Learning (ICML)},
  year={2011}
}

@misc{zhong2025accurateposttrainingquantizationvision,
      title={Towards Accurate Post-Training Quantization of Vision Transformers via Error Reduction}, 
      author={Yunshan Zhong and You Huang and Jiawei Hu and Yuxin Zhang and Rongrong Ji},
      year={2025},
      eprint={2407.06794},
      archivePrefix={arXiv},
      primaryClass={cs.CV},
      url={https://arxiv.org/abs/2407.06794}, 
}

@misc{frantar2023gptqaccurateposttrainingquantization,
      title={GPTQ: Accurate Post-Training Quantization for Generative Pre-trained Transformers}, 
      author={Elias Frantar and Saleh Ashkboos and Torsten Hoefler and Dan Alistarh},
      year={2023},
      eprint={2210.17323},
      archivePrefix={arXiv},
      primaryClass={cs.LG},
      url={https://arxiv.org/abs/2210.17323}, 
}

@inproceedings{jakubovitz2018jacobian,
  title={Improving {DNN} Robustness to Adversarial Attacks using Jacobian Regularization},
  author={Jakubovitz, Daniel and Giryes, Raja},
  booktitle={European Conference on Computer Vision (ECCV)},
  year={2018}
}

@article{nguyen2024discriminative,
  title={Training More Robust Classification Model via Discriminative Loss and Gaussian Noise Injection},
  author={Nguyen, Hoang-Viet and Gamboa, Fabrice and Zhang, Sixin and Chhaibi, Reda and Gratton, Serge and Giaccone, Thomas},
  journal={Transactions on Machine Learning Research},
  year={2024}
}

@inproceedings{chen2017noisy,
  title={Noisy Softmax: Improving the Generalization Ability of {DCNN} via Postponing the Early Softmax Saturation},
  author={Chen, Binghui and Deng, Weihong and Du, Junping},
  booktitle={IEEE Conference on Computer Vision and Pattern Recognition (CVPR)},
  year={2017}
}

@article{bishop1995training,
  title={Training with Noise is Equivalent to {T}ikhonov Regularization},
  author={Bishop, Christopher M},
  journal={Neural Computation},
  volume={7},
  number={1},
  pages={108--116},
  year={1995}
}

@inproceedings{frantar2023gptq,
  title={{GPTQ}: Accurate Post-Training Quantization for Generative Pre-trained Transformers},
  author={Frantar, Elias and Ashkboos, Saleh and Hoefler, Torsten and Alistarh, Dan},
  booktitle={International Conference on Learning Representations (ICLR)},
  year={2023}
}

@inproceedings{lin2024awq,
  title={{AWQ}: Activation-aware Weight Quantization for {LLM} Compression and Acceleration},
  author={Lin, Ji and Tang, Jiaming and Tang, Haotian and Yang, Shang and Dang, Xingyu and Han, Song},
  booktitle={Conference on Machine Learning and Systems (MLSys)},
  note={Best Paper Award},
  year={2024}
}

@inproceedings{xiao2023smoothquant,
  title={{SmoothQuant}: Accurate and Efficient Post-Training Quantization for Large Language Models},
  author={Xiao, Guangxuan and Lin, Ji and Seznec, Mickael and Wu, Hao and Demouth, Julien and Han, Song},
  booktitle={International Conference on Machine Learning (ICML)},
  year={2023}
}

@inproceedings{dettmers2022llmint8,
  title={{LLM}.int8(): 8-bit Matrix Multiplication for Transformers at Scale},
  author={Dettmers, Tim and Lewis, Mike and Belkada, Younes and Zettlemoyer, Luke},
  booktitle={Advances in Neural Information Processing Systems (NeurIPS)},
  year={2022}
}

@inproceedings{esser2020learned,
  title={Learned Step Size Quantization},
  author={Esser, Steven K and McKinstry, Jeffrey L and Bablani, Deepika and Appuswamy, Rathinakumar and Modha, Dharmendra S},
  booktitle={International Conference on Learning Representations (ICLR)},
  year={2020}
}

@article{choi2018pact,
  title={{PACT}: Parameterized Clipping Activation for Quantized Neural Networks},
  author={Choi, Jungwook and Wang, Zhuo and Venkataramani, Swagath and Chuang, Pierce I and Srinivasan, Vijayalakshmi and Gopalakrishnan, Kailash},
  journal={arXiv preprint arXiv:1805.06085},
  year={2018}
}

@inproceedings{jacob2018quantization,
  title={Quantization and Training of Neural Networks for Efficient Integer-Arithmetic-Only Inference},
  author={Jacob, Benoit and Kligys, Skirmantas and Chen, Bo and Zhu, Menglong and Tang, Matthew and Howard, Andrew and Adam, Hartwig and Kalenichenko, Dmitry},
  booktitle={IEEE Conference on Computer Vision and Pattern Recognition (CVPR)},
  year={2018}
}

@inproceedings{kim2021ibert,
  title={{I-BERT}: Integer-only {BERT} Quantization},
  author={Kim, Sehoon and Gholami, Amir and Yao, Zhewei and Mahoney, Michael W and Keutzer, Kurt},
  booktitle={International Conference on Machine Learning (ICML)},
  year={2021}
}

@inproceedings{lin2022fqvit,
  title={{FQ-ViT}: Post-Training Quantization for Fully Quantized Vision Transformer},
  author={Lin, Yang and Zhang, Tianyu and Sun, Peiqin and Li, Zheng and Zhou, Shuchang},
  booktitle={International Joint Conference on Artificial Intelligence (IJCAI)},
  year={2022}
}

@inproceedings{li2023repqvit,
  title={{RepQ-ViT}: Scale Reparameterization for Post-Training Quantization of Vision Transformers},
  author={Li, Yuhang and Gong, Ruihao and Tan, Xu and Yang, Yang and Hu, Peng and Zhang, Qi and Yu, Fengwei and Wang, Wei and Gu, Shi},
  booktitle={IEEE International Conference on Computer Vision (ICCV)},
  year={2023}
}

@inproceedings{liu2021posttraining,
  title={Post-Training Quantization for Vision Transformer},
  author={Liu, Zhenhua and Wang, Yunhe and Han, Kai and Zhang, Wei and Ma, Siwei and Gao, Wen},
  booktitle={Advances in Neural Information Processing Systems (NeurIPS)},
  year={2021}
}

@inproceedings{vaswani2017attention,
  title={Attention is All You Need},
  author={Vaswani, Ashish and Shazeer, Noam and Parmar, Niki and Uszkoreit, Jakob and Jones, Llion and Gomez, Aidan N and Kaiser, Lukasz and Polosukhin, Illia},
  booktitle={Advances in Neural Information Processing Systems (NeurIPS)},
  year={2017}
}

@article{dosovitskiy2021image,
  title={An Image is Worth 16x16 Words: Transformers for Image Recognition at Scale},
  author={Dosovitskiy, Alexey and Beyer, Lucas and Kolesnikov, Alexander and Weissenborn, Dirk and Zhai, Xiaohua and Unterthiner, Thomas and Dehghani, Mostafa and Minderer, Matthias and Heigold, Georg and Gelly, Sylvain and Uszkoreit, Jakob and Houlsby, Neil},
  journal={International Conference on Learning Representations (ICLR)},
  year={2021}
}

@article{touvron2023llama,
  title={{LLaMA}: Open and Efficient Foundation Language Models},
  author={Touvron, Hugo and Lavril, Thibaut and Izacard, Gautier and Martinet, Xavier and Lachaux, Marie-Anne and Lacroix, Timoth{\'e}e and Rozi{\`e}re, Baptiste and Goyal, Naman and Hambro, Eric and Azhar, Faisal and others},
  journal={arXiv preprint arXiv:2302.13971},
  year={2023}
}

@article{touvron2023llama2,
  title={{LLaMA} 2: Open Foundation and Fine-Tuned Chat Models},
  author={Touvron, Hugo and Martin, Louis and Stone, Kevin and Albert, Peter and Almahairi, Amjad and Babaei, Yasmine and Bashlykov, Nikolay and Batra, Soumya and Bhargava, Prajjwal and Bhosale, Shruti and others},
  journal={arXiv preprint arXiv:2307.09288},
  year={2023}
}

@article{grattafiori2024llama3,
  title={The {LLaMA} 3 Herd of Models},
  author={Grattafiori, Aaron and Dubey, Abhimanyu and Jauhri, Abhinav and others},
  journal={arXiv preprint arXiv:2407.21783},
  year={2024}
}

@article{qwen2024qwen25,
  title={{Qwen2.5} Technical Report},
  author={Qwen Team},
  journal={arXiv preprint arXiv:2412.15115},
  year={2024}
}

@article{zhai2023siglip,
  title={Sigmoid Loss for Language Image Pre-Training},
  author={Zhai, Xiaohua and Mustafa, Basil and Kolesnikov, Alexander and Beyer, Lucas},
  journal={IEEE International Conference on Computer Vision (ICCV)},
  year={2023}
}

@article{srivastava2014dropout,
  title={Dropout: A Simple Way to Prevent Neural Networks from Overfitting},
  author={Srivastava, Nitish and Hinton, Geoffrey and Krizhevsky, Alex and Sutskever, Ilya and Salakhutdinov, Ruslan},
  journal={Journal of Machine Learning Research},
  volume={15},
  number={1},
  pages={1929--1958},
  year={2014}
}

@inproceedings{wan2013regularization,
  title={Regularization of Neural Networks using {DropConnect}},
  author={Wan, Li and Zeiler, Matthew and Zhang, Sixin and Le Cun, Yann and Fergus, Rob},
  booktitle={International Conference on Machine Learning (ICML)},
  year={2013}
}

@inproceedings{goodfellow2015explaining,
  title={Explaining and Harnessing Adversarial Examples},
  author={Goodfellow, Ian J and Shlens, Jonathon and Szegedy, Christian},
  booktitle={International Conference on Learning Representations (ICLR)},
  year={2015}
}

@inproceedings{hendrycks2019benchmarking,
  title={Benchmarking Neural Network Robustness to Common Corruptions and Perturbations},
  author={Hendrycks, Dan and Dietterich, Thomas},
  booktitle={International Conference on Learning Representations (ICLR)},
  year={2019}
}

@inproceedings{madry2018towards,
  title={Towards Deep Learning Models Resistant to Adversarial Attacks},
  author={Madry, Aleksander and Makelov, Aleksandar and Schmidt, Ludwig and Tsipras, Dimitris and Vladu, Adrian},
  booktitle={International Conference on Learning Representations (ICLR)},
  year={2018}
}

@inproceedings{micikevicius2018mixed,
  title={Mixed Precision Training},
  author={Micikevicius, Paulius and Narang, Sharan and Alben, Jonah and Diamos, Gregory and Elsen, Erich and Garcia, David and Ginsburg, Boris and Houston, Michael and Kuchaiev, Oleksii and Venkatesh, Ganesh and Wu, Hao},
  booktitle={International Conference on Learning Representations (ICLR)},
  year={2018}
}

@article{hendrycks2021mmlu,
  title={Measuring Massive Multitask Language Understanding},
  author={Hendrycks, Dan and Burns, Collin and Basart, Steven and Zou, Andy and Mazeika, Mantas and Song, Dawn and Steinhardt, Jacob},
  journal={International Conference on Learning Representations (ICLR)},
  year={2021}
}

@article{zellers2019hellaswag,
  title={{HellaSwag}: Can a Machine Really Finish Your Sentence?},
  author={Zellers, Rowan and Holtzman, Ari and Bisk, Yonatan and Farhadi, Ali and Choi, Yejin},
  journal={Association for Computational Linguistics (ACL)},
  year={2019}
}

@article{sakaguchi2020winogrande,
  title={{WinoGrande}: An Adversarial Winograd Schema Challenge at Scale},
  author={Sakaguchi, Keisuke and Bras, Ronan Le and Bhagavatula, Chandra and Choi, Yejin},
  journal={AAAI Conference on Artificial Intelligence},
  year={2020}
}

@article{bisk2020piqa,
  title={{PIQA}: Reasoning about Physical Commonsense in Natural Language},
  author={Bisk, Yonatan and Zellers, Rowan and Bras, Ronan Le and Gao, Jianfeng and Choi, Yejin},
  journal={AAAI Conference on Artificial Intelligence},
  year={2020}
}

@article{lin2022truthfulqa,
  title={{TruthfulQA}: Measuring How Models Mimic Human Falsehoods},
  author={Lin, Stephanie and Hilton, Jacob and Evans, Owain},
  journal={Association for Computational Linguistics (ACL)},
  year={2022}
}

@article{merity2017pointer,
  title={Pointer Sentinel Mixture Models},
  author={Merity, Stephen and Xiong, Caiming and Bradbury, James and Socher, Richard},
  journal={International Conference on Learning Representations (ICLR)},
  year={2017}
}

@inproceedings{deng2009imagenet,
  title={{ImageNet}: A Large-Scale Hierarchical Image Database},
  author={Deng, Jia and Dong, Wei and Socher, Richard and Li, Li-Jia and Li, Kai and Fei-Fei, Li},
  booktitle={IEEE Conference on Computer Vision and Pattern Recognition (CVPR)},
  year={2009}
}

@article{tang2024survey,
  title={A Survey on Transformer Compression},
  author={Tang, Yehui and Wang, Yunhe and Guo, Jianyuan and Tu, Zhijun and Han, Kai and Hu, Hailin and Tao, Dacheng},
  journal={arXiv preprint arXiv:2402.05964},
  year={2024}
}

@article{gholami2022survey,
  title={A Survey of Quantization Methods for Efficient Neural Network Inference},
  author={Gholami, Amir and Kim, Sehoon and Dong, Zhen and Yao, Zhewei and Mahoney, Michael W and Keutzer, Kurt},
  journal={Low-Power Computer Vision},
  year={2022}
}

@article{taori2023alpaca,
  title={Stanford {Alpaca}: An Instruction-following {LLaMA} Model},
  author={Taori, Rohan and Gulrajani, Ishaan and Zhang, Tianyi and Dubois, Yann and Li, Xuechen and Guestrin, Carlos and Liang, Percy and Hashimoto, Tatsunori B},
  journal={GitHub repository},
  year={2023}
}

@inproceedings{guo2017calibration,
  title={On Calibration of Modern Neural Networks},
  author={Guo, Chuan and Pleiss, Geoff and Sun, Yu and Weinberger, Kilian Q},
  booktitle={International Conference on Machine Learning (ICML)},
  year={2017}
}

@article{loshchilov2019decoupled,
  title={Decoupled Weight Decay Regularization},
  author={Loshchilov, Ilya and Hutter, Frank},
  journal={International Conference on Learning Representations (ICLR)},
  year={2019}
}

@inproceedings{dong2019hawq,
  title={{HAWQ}: Hessian {AW}are Quantization of Neural Networks with Mixed-Precision},
  author={Dong, Zhen and Yao, Zhewei and Gholami, Amir and Mahoney, Michael W and Keutzer, Kurt},
  booktitle={IEEE International Conference on Computer Vision (ICCV)},
  year={2019}
}

@inproceedings{wang2019haq,
  title={{HAQ}: Hardware-Aware Automated Quantization with Mixed Precision},
  author={Wang, Kuan and Liu, Zhijian and Lin, Yujun and Lin, Ji and Han, Song},
  booktitle={IEEE Conference on Computer Vision and Pattern Recognition (CVPR)},
  year={2019}
}
\bibliographystyle{icml2026}

\newpage
\appendix
\onecolumn

\section{Derivation of Closed-Form Jacobian Frobenius Norm}
\label{app:derivation}

For completeness, we derive the closed-form expression for $||J_S||_F^2$.

The Softmax Jacobian is $J_S = \text{diag}(p) - pp^T$, where $p = S(z)$. The Frobenius norm squared is:
\begin{align}
    ||J_S||_F^2 &= \text{Tr}(J_S^T J_S) \\
    &= \text{Tr}\left((\text{diag}(p) - pp^T)^T (\text{diag}(p) - pp^T)\right) \\
    &= \text{Tr}\left(\text{diag}(p)^2 - 2\text{diag}(p)pp^T + pp^Tpp^T\right) \\
    &= \sum_i p_i^2 - 2\sum_i p_i \cdot p_i \cdot \sum_j p_j + \left(\sum_i p_i^2\right)^2 \\
    &= ||p||_2^2 - 2||p||_3^3 + ||p||_2^4
\end{align}

where we used $\sum_j p_j = 1$ (softmax normalization) and $\text{Tr}(pp^Tpp^T) = (p^Tp)^2 = ||p||_2^4$.

\section{Experimental Details}
\label{app:experimental_details}

This section provides comprehensive details on training configurations, quantization strategies, and infrastructure used in our experiments.

\subsection{Training Configuration}

\textbf{Optimizer and Learning Rate.} We use AdamW~\cite{loshchilov2019decoupled} with learning rate $5 \times 10^{-7}$, weight decay $0.01$, and gradient clipping at max norm $1.0$. Training uses a cosine learning rate scheduler with 500 warmup steps.

\textbf{Batch Size and Accumulation.} We use per-device batch size of 1 with gradient accumulation over 8 steps, yielding an effective batch size of 8 per GPU. For multi-GPU training, the effective batch size scales with the number of GPUs.

\textbf{Training Duration.} Models are fine-tuned for 5 epochs on the Alpaca dataset~\cite{taori2023alpaca} with maximum sequence length of 2048 tokens. Evaluation is performed every 2000 steps.

\textbf{Precision.} All training uses bfloat16 mixed precision for memory efficiency and numerical stability.

\subsection{Jacobian-Guided Noise Injection Parameters}

\textbf{Noise Scaling.} The noise scaling coefficient $\alpha$ is set to 0.5 (see \cref{fig:jacobian_norm}(b) for sensitivity analysis). Noise standard deviation is clamped to $[\sigma_{\min}, \sigma_{\max}] = [0.005, 10.0]$ for numerical stability.

\textbf{Amortized Updates.} To reduce computational overhead, Jacobian norms are computed and noise parameters are updated every $T = 100$ training steps rather than every step. This amortization has negligible impact on final performance while reducing overhead.

\textbf{Warmup.} Jacobian-guided noise updates begin after 100 warmup steps to allow initial model stabilization. During warmup, a constant base noise with $\sigma = 0.1$ is applied.

\textbf{Noise Mode.} We use rowwise (per-position) noise injection where each query position receives noise calibrated to its local Jacobian norm. This is more effective than layerwise or global noise as it targets high-sensitivity positions specifically.

\subsection{Post-Training Quantization Methods}

\textbf{AWQ}~\cite{lin2024awq}: Activation-aware Weight Quantization identifies salient weights based on activation magnitudes and applies per-channel scaling to protect these weights during quantization. We use the default calibration set of 128 samples.

\textbf{GPTQ}~\cite{frantar2023gptq}: Uses approximate second-order information (Hessian) to quantize weights layer-by-layer while minimizing reconstruction error. We use group size of 128 for weight quantization.

\textbf{SpinQuant}~\cite{liu2025spinquantllmquantizationlearned}: A PTQ method that applies learned orthogonal (rotation) transformations to weights and activations to reduce outliers and make distributions more uniform before
  quantization, also supporting KV cache quantization.

\subsection{Quantization Bit-Width Configurations}

We evaluate multiple quantization configurations denoted as WxAy (x-bit weights, y-bit activations):

\begin{table}[h]
\centering
\begin{tabular}{lcc}
\toprule
\textbf{Config} & \textbf{Weight Bits} & \textbf{Activation Bits} \\
\midrule
W4A4 & 4 & 4 \\
W4A8 & 4 & 8 \\
W5A5 & 5 & 5 \\
W6A6 & 6 & 6 \\
W6A8 & 6 & 8 \\
W8A8 & 8 & 8 \\
\bottomrule
\end{tabular}
\end{table}

Weight quantization uses per-channel granularity. Activation quantization uses per-token granularity for language models.

\subsection{Distributed Training Infrastructure}

\textbf{DeepSpeed Configuration.} We use DeepSpeed ZeRO Stage 2 for memory-efficient distributed training with the following settings:
\begin{itemize}
    \item Gradient partitioning with allgather bucket size $2 \times 10^8$
    \item Reduce scatter with bucket size $2 \times 10^8$
    \item Communication overlap enabled
    \item Contiguous gradient buffers
\end{itemize}

\textbf{Hardware.} Experiments are conducted on 8 NVIDIA A100 GPUs with 80 GB memory each. We train both Llama and Qwen models for up to 10 epochs until convergence.

\subsection{Evaluation Protocol}

\textbf{Language Model Benchmarks.} We evaluate in a zero shot setting on HellaSwag, Winogrande, PIQA, TruthfulQA, WikiText, BoolQ, Arc-Easy and Arc-Challenge.

\textbf{Vision Model Benchmarks.} SigLIP models are evaluated on ImageNet-1K~\cite{deng2009imagenet} zero-shot classification using the standard CLIP evaluation protocol with Top-1 and Top-5 accuracy metrics.

\section{SigLIP Post-Training Quantization Results}
\label{app:siglip_ptq}

\cref{tab:siglip_results_full} provides detailed results for SigLIP base 16-384 under post-training quantization with ERQ and RepQViT methods.

\textbf{Model Configuration.} We use the SigLIP base 16-384 vision encoder~\cite{zhai2023siglip}, which processes images at 384$\times$384 resolution with 16$\times$16 patches. The model is evaluated on ImageNet-1K~\cite{deng2009imagenet} classification using the partitioned test set (split 95-5).

\textbf{Quantization Methods.} We evaluate two PTQ methods designed for vision transformers:
\begin{itemize}
    \item \textbf{ERQ} (Error-aware Quantization): A PTQ method that minimizes reconstruction error by considering the error propagation through transformer layers. ERQ optimizes quantization parameters to reduce the cumulative error in attention and feed-forward computations.
    \item \textbf{RepQViT}~\cite{li2023repqvit}: A reparameterization-based approach that decouples quantization scales for hardware-friendly deployment. RepQViT addresses the unique challenges of quantizing vision transformers by handling post-LayerNorm activations and attention score distributions.
\end{itemize}

\textbf{Bit-width Settings.} We evaluate four quantization configurations denoted as WxAy, where x is the weight bit-width and y is the activation bit-width:
\begin{itemize}
    \item \textbf{W4A4}: Aggressive 4-bit quantization for both weights and activations, suitable for edge deployment with severe memory constraints.
    \item \textbf{W5A5}: Moderate 5-bit quantization offering a balance between compression and accuracy.
    \item \textbf{W6A6}: Conservative 6-bit quantization with minimal accuracy degradation.
    \item \textbf{W8A8}: Near-lossless 8-bit quantization serving as a reference point.
\end{itemize}

\textbf{Training Protocol.} Models are fine-tuned with Jacobian-guided noise injection as described in \cref{sec:methodology}, using the hyperparameters specified in \cref{sec:experiments}. The noise scaling coefficient $\alpha$ is set to 0.5 based on the sensitivity analysis in \cref{fig:jacobian_norm}.

\begin{table*}[t]
\centering
\caption{SigLIP base 16-384 PTQ results on ImageNet-1K zero-shot classification. All values are Top-1 accuracy (\%). B: Baseline, B+N: Baseline with Jacobian-guided Noise. The FP32 baseline achieves 86.7\% Top-1 accuracy.}
\label{tab:siglip_results_full}
\begin{tabular}{llcccc}
\toprule
\textbf{Method} & \textbf{Noise} & \textbf{W4A4} & \textbf{W5A5} & \textbf{W6A6} & \textbf{W8A8} \\
\midrule
\multirow{2}{*}{ERQ} & B & 51.2 & 72.4 & 81.3 & 85.9 \\
 & B+N & \textbf{60.1} & \textbf{76.8} & \textbf{83.5} & \textbf{86.4} \\
\midrule
\multirow{2}{*}{RepQViT} & B & 32.5 & 65.1 & 78.9 & 85.6 \\
 & B+N & \textbf{69.8} & \textbf{74.2} & \textbf{82.1} & \textbf{86.2} \\
\bottomrule
\end{tabular}
\end{table*}

\section{Full-Precision Results With and Without Noise}
\label{app:fp16_noise}

\cref{tab:qwen_fp16_noise} presents evaluation results for Qwen2.5-3B at full precision (FP16) with and without Jacobian-guided noise injection during training. This demonstrates that noise injection does not degrade full-precision model performance while providing quantization robustness.

\begin{table*}[t]
\centering
\caption{Qwen2.5-3B full-precision (FP16) results with and without noise injection. B: Baseline, B+N: With Jacobian Noise ($\alpha=0.5$, rowwise).}
\label{tab:qwen_fp16_noise}
\begin{tabular}{lcccccc}
\toprule
\textbf{Model} & \textbf{MMLU} & \textbf{HellaSwag} & \textbf{PIQA} & \textbf{Winogrande} & \textbf{TruthfulQA} & \textbf{WikiText PPL} \\
\midrule
Qwen2.5-3B (B) & 65.25 & 75.08 & 79.87 & 70.72 & 47.41 & 10.60 \\
Qwen2.5-3B (B+N) & 65.48 & 74.91 & 79.54 & 70.17 & 47.47 & 10.46 \\
\bottomrule
\end{tabular}
\end{table*}

\section{Detailed zero-shot PTQ results for Llama and Qwen}

Tables \ref{tab:spinquant-bitwidth-sweep},  \ref{tab:awq-bitwidth-sweep} and  \ref{tab:gptq-bitwidth-sweep} shows the detailed results for Llama and Qwen model on all 7 benchmarks. We observe that noise injections improves the performance of the overall model, while maintaining comparable or better performance across all 7 benchmarks.

\begin{table}[h]
\centering
\small
\caption{SpinQuant W4A4KV4 / W4A8KV8 / W6A6KV6 / W6A8KV8 full results for Qwen2.5-3B and Llama-3.2-3B on both the Base and Jacobian-noise (Jac) variants. PPL is WikiText-2. Zero-shot metric is \texttt{acc\_norm} for HellaSwag/PIQA/ARC-C and \texttt{acc} otherwise. ``Avg7'' is the average of HellaSwag, PIQA, WinoGrande, ARC-E, ARC-C, BoolQ, TQA-MC2. }
\label{tab:spinquant-bitwidth-sweep}
\resizebox{\textwidth}{!}{%
\begin{tabular}{lll*{9}{r}}
\toprule
Model & Variant & Precision & PPL & HellaSwag & PIQA & WinoGrande & ARC-E & ARC-C & BoolQ & TQA-MC2 & Avg7 \\
\midrule

\multirow{10}{*}{Qwen2.5-3B}
  & \multirow{4}{*}{Base}
  & W4A4KV4  & 10.49 & 66.27 & 74.32 & 59.19 & 71.93 & 44.20 & 65.32 & 41.12 & 60.34 \\
  & & W4A8KV8  &  8.55 & 72.91 & 78.02 & 68.98 & 77.48 & 48.21 & 79.94 & 45.91 & 67.35 \\
  & & W6A6KV6  &  8.36 & 73.92 & 78.35 & 67.25 & 77.15 & 48.38 & 79.08 & 46.64 & 67.25 \\
  & & W6A8KV8  &  8.32 & 74.25 & 79.33 & 68.59 & 76.18 & 48.98 & 79.57 & 46.74 & 67.66 \\
\cmidrule(l){2-12}
  & \multirow{4}{*}{Jac-noise}
  & W4A4KV4  & 10.33 & 66.65 & 74.43 & 61.88 & 73.70 & 45.05 & 70.18 & 46.42 & 62.62 \\
  & & W4A8KV8  &  8.35 & 72.32 & 78.40 & 68.35 & 74.03 & 45.73 & 77.55 & 43.25 & 65.66 \\
  & & W6A6KV6  &  8.18 & 73.26 & 78.62 & 68.27 & 77.69 & 48.81 & 79.08 & 46.32 & 67.44 \\
  & & W6A8KV8  &  8.11 & 73.56 & 78.89 & 67.64 & 77.10 & 49.40 & 79.54 & 46.03 & 67.45 \\
\midrule

\multirow{10}{*}{Llama-3.2-3B}
  & \multirow{4}{*}{Base}
  & W4A4KV4  & 10.72 & 66.96 & 74.05 & 60.38 & 69.99 & 39.76 & 71.71 & 38.89 & 60.25 \\
  & & W4A8KV8  &  8.31 & 72.44 & 77.37 & 68.82 & 73.65 & 43.77 & 74.16 & 41.12 & 64.48 \\
  & & W6A6KV6  &  8.03 & 73.74 & 78.67 & 69.46 & 76.47 & 46.59 & 75.29 & 41.66 & 65.98 \\
  & & W6A8KV8  &  7.97 & 73.86 & 78.13 & 69.93 & 74.07 & 47.78 & 76.24 & 41.64 & 65.95 \\
\cmidrule(l){2-12}
  & \multirow{4}{*}{Jac-noise}
  & W4A4KV4  & 11.18 & 67.38 & 73.50 & 64.09 & 67.05 & 40.27 & 70.40 & 38.35 & 60.15 \\
  & & W4A8KV8  &  8.47 & 73.33 & 77.97 & 68.35 & 74.71 & 46.33 & 73.85 & 41.65 & 65.17 \\
  & & W6A6KV6  &  8.17 & 74.67 & 77.91 & 70.88 & 77.31 & 48.38 & 76.24 & 42.31 & 66.81 \\
  & & W6A8KV8  &  8.12 & 74.94 & 78.35 & 70.01 & 75.13 & 49.66 & 76.27 & 42.70 & 66.72 \\
\bottomrule
\end{tabular}%
}
\end{table}

\begin{table}[h]
\centering
\small
\caption{AWQ W4A4 / W4A8 / W6A6 / W8A8 results for Qwen2.5-3B and Llama-3.2-3B on both the Base and Jacobian-noise (Jac) variants. PPL is WikiText-2. Zero-shot metric is \texttt{acc\_norm} for HellaSwag/PIQA/ARC-C and \texttt{acc} otherwise. ``Avg7'' is the average of HellaSwag, PIQA, WinoGrande, ARC-E, ARC-C, BoolQ, TQA-MC2.}
\label{tab:awq-bitwidth-sweep}
\resizebox{\textwidth}{!}{%
\begin{tabular}{lll*{9}{r}}
\toprule
Model & Variant & Precision & PPL & HellaSwag & PIQA & WinoGrande & ARC-E & ARC-C & BoolQ & TQA-MC2 & Avg7 \\
\midrule

\multirow{8}{*}{Qwen2.5-3B}
  & \multirow{4}{*}{Base}
  & W4A4     & 8121.24 & 30.23 & 54.35 & 48.54 & 32.32 & 23.12 & 48.07 & 47.45 & 40.58 \\
  & & W4A8     &   10.80 & 74.23 & 77.97 & 68.11 & 78.32 & 49.57 & 80.18 & 46.21 & 67.80 \\
  & & W6A6     &   15.29 & 70.26 & 74.86 & 64.72 & 75.80 & 49.40 & 75.54 & 45.25 & 65.12 \\
  & & W8A8     &   10.80 & 74.23 & 77.80 & 68.59 & 77.99 & 50.00 & 80.70 & 45.57 & 67.84 \\
\cmidrule(l){2-12}
  & \multirow{4}{*}{Jac-noise }
  & W4A4     & 3281.04 & 30.57 & 53.26 & 52.25 & 33.54 & 24.06 & 49.42 & 47.80 & 41.56 \\
  & & W4A8     &   11.01 & 73.45 & 78.24 & 68.51 & 78.11 & 49.40 & 80.03 & 44.61 & 67.48 \\
  & & W6A6     &   13.63 & 70.95 & 75.81 & 64.61 & 76.29 & 48.81 & 74.84 & 46.35 & 65.38 \\
  & & W8A8     &   10.80 & 74.44 & 79.40 & 68.25 & 79.20 & 50.49 & 81.12 & 45.73 & 68.38 \\
\midrule

\multirow{8}{*}{Llama-3.2-3B}
  & \multirow{4}{*}{Base}
  & W4A4     &  105.06 & 32.47 & 55.17 & 51.14 & 33.71 & 23.81 & 50.80 & 47.31 & 42.06 \\
  & & W4A8     &    9.96 & 74.58 & 78.24 & 71.03 & 77.06 & 47.61 & 76.39 & 41.64 & 66.65 \\
  & & W6A6     &   10.80 & 72.87 & 76.99 & 68.35 & 74.45 & 45.05 & 70.58 & 40.21 & 64.07 \\
  & & W8A8     &    9.73 & 74.42 & 78.62 & 70.96 & 77.10 & 47.95 & 76.27 & 41.04 & 66.62 \\
\cmidrule(l){2-12}
  & \multirow{4}{*}{Jac-noise}
  & W4A4     &  104.34 & 33.33 & 55.41 & 51.04 & 33.79 & 24.29 & 51.55 & 45.99 & 42.20 \\
  & & W4A8     &    9.86 & 75.17 & 78.78 & 70.48 & 78.24 & 49.66 & 77.13 & 42.00 & 67.35 \\
  & & W6A6     &   10.55 & 73.46 & 77.04 & 67.88 & 75.08 & 47.01 & 72.97 & 40.52 & 64.85 \\
  & & W8A8     &    9.68 & 75.41 & 78.94 & 70.72 & 78.54 & 49.74 & 76.91 & 42.61 & 67.55 \\
\bottomrule
\end{tabular}%
}
\end{table}

\begin{table}[h]
\centering
\small
\caption{GPTQ W4A4 / W4A8 / W6A6 / W8A8 results for Qwen2.5-3B and Llama-3.2-3B on both the Base and Jacobian-noise (Jac) variants. PPL is WikiText-2. Zero-shot metric is \texttt{acc\_norm} for HellaSwag/PIQA/ARC-C and \texttt{acc} otherwise. ``Avg7'' is the average of HellaSwag, PIQA, WinoGrande, ARC-E, ARC-C, BoolQ, TQA-MC2.}
\label{tab:gptq-bitwidth-sweep}
\resizebox{\textwidth}{!}{%
\begin{tabular}{lll*{9}{r}}
\toprule
Model & Variant & Precision & PPL & HellaSwag & PIQA & WinoGrande & ARC-E & ARC-C & BoolQ & TQA-MC2 & Avg7 \\
\midrule

\multirow{8}{*}{Qwen2.5-3B}
  & \multirow{4}{*}{Base}
    & W4A4     & 5172.39 & 25.81 & 49.95 & 51.14 & 26.52 & 23.46 & 43.91 & 49.27 & 38.58 \\
  & & W4A8     &   14.13 & 72.05 & 77.15 & 65.98 & 78.00 & 48.89 & 80.03 & 44.43 & 66.65 \\
  & & W6A6     &   28.42 & 63.32 & 70.78 & 60.46 & 69.15 & 42.66 & 68.23 & 45.19 & 59.97 \\
  & & W8A8     &   10.98 & 73.99 & 78.02 & 67.88 & 77.99 & 49.15 & 80.06 & 45.78 & 67.55 \\
\cmidrule(l){2-12}
  & \multirow{4}{*}{Jac-noise}
    & W4A4     & 4207.85 & 26.31 & 50.05 & 51.04 & 26.00 & 24.32 & 45.38 & 49.64 & 38.96 \\
  & & W4A8     &   12.21 & 71.39 & 77.42 & 66.77 & 77.23 & 49.40 & 77.43 & 43.14 & 66.11 \\
  & & W6A6     &   21.52 & 63.10 & 71.98 & 60.77 & 69.28 & 43.86 & 69.20 & 42.54 & 60.10 \\
  & & W8A8     &   10.82 & 73.41 & 78.65 & 68.23 & 78.95 & 48.91 & 80.36 & 44.73 & 67.61 \\
\midrule

\multirow{8}{*}{Llama-3.2-3B}
  & \multirow{4}{*}{Base}
    & W4A4     & 454.35 & 29.75 & 51.36 & 49.88 & 30.89 & 23.55 & 46.79 & 49.96 & 40.31 \\
  & & W4A8     &   37.40 & 72.75 & 77.91 & 67.80 & 74.20 & 44.97 & 76.91 & 39.10 & 64.81 \\
  & & W6A6     &   10.87 & 72.05 & 76.55 & 66.61 & 73.32 & 45.31 & 71.53 & 40.49 & 63.69 \\
  & & W8A8     &    9.69 & 74.37 & 78.62 & 71.27 & 76.77 & 48.38 & 75.93 & 41.50 & 66.69 \\
\cmidrule(l){2-12}
  & \multirow{4}{*}{Jac-noise}
    & W4A4     & 496.38 & 30.84 & 54.21 & 48.67 & 30.76 & 25.91 & 46.87 & 47.77 & 41.00 \\
  & & W4A8     &   31.47 & 73.53 & 78.13 & 70.96 & 76.85 & 47.61 & 76.64 & 40.82 & 66.36 \\
  & & W6A6     &   11.12 & 73.19 & 76.66 & 66.46 & 74.96 & 45.48 & 71.56 & 40.81 & 64.16 \\
  & & W8A8     &    9.86 & 75.21 & 78.56 & 69.53 & 78.07 & 49.40 & 77.37 & 42.64 & 67.25 \\
\bottomrule
\end{tabular}%
}
\end{table}

\section{Activation Similarity Analysis on Hardware}
\label{app:production_similarity}

To validate our method in real-world deployment scenarios, we analyze a Vision-Language Model (VLM) trained with quantization-aware training at 15-bit precision and deployed on edge hardware. We measure the cosine similarity between quantized and unquantized activations at each layer, providing a direct measure of how well the quantized model preserves the computational behavior of the full-precision model.

\begin{figure*}[h]
\centering
\begin{subfigure}[b]{\linewidth}
    \includegraphics[width=\linewidth]{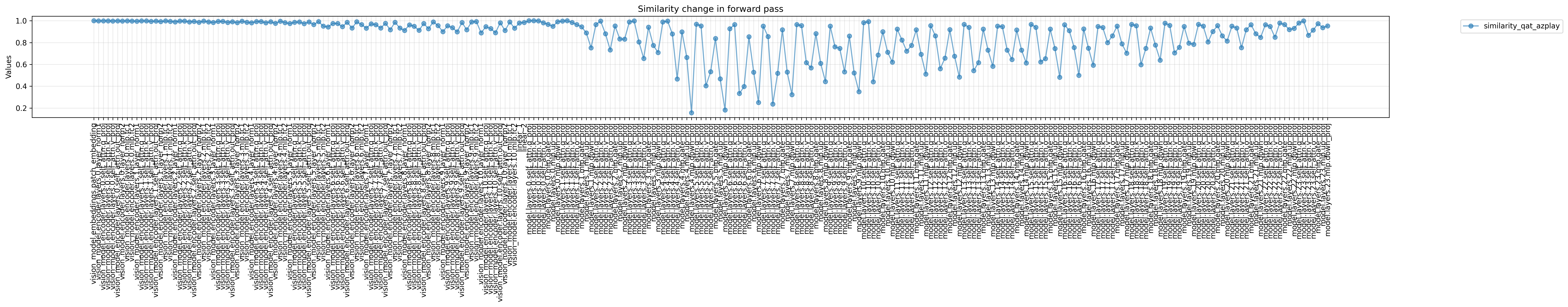}
    \caption{}
    \label{fig:production_similarity_base}
\end{subfigure}
\vspace{0.5em}
\begin{subfigure}[b]{\linewidth}
    \includegraphics[width=\linewidth]{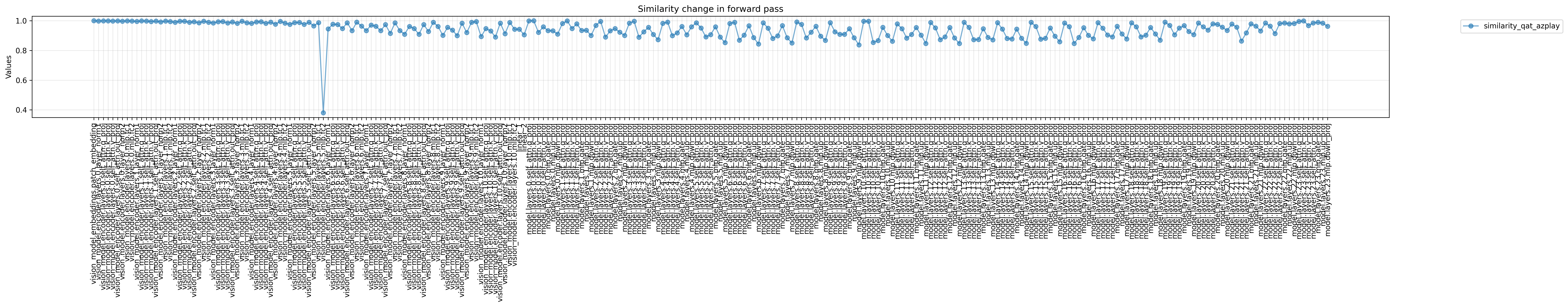}
    \caption{}
    \label{fig:production_similarity_noise}
\end{subfigure}
\caption{Layer-wise cosine similarity between quantized (15-bit) and unquantized activations for a VLM deployed on edge hardware. (a) Baseline model shows severe similarity degradation, with some layers dropping to near-zero cosine similarity. (b) Model trained with Jacobian-guided noise injection maintains consistently high similarity ($>$0.95) across all layers.}
\label{fig:production_similarity}
\end{figure*}

\textbf{Experimental Setup.} The VLM is deployed on edge accelerator hardware with 15-bit fixed-point arithmetic. Unlike standard GPU deployment where numerical behavior closely matches training (bfloat16/float32), edge accelerators often have distinct rounding behavior that can cause divergent generation even at relatively high bit-widths.

\textbf{Results.} Figure \ref{fig:production_similarity} reveals a striking difference between baseline and noise-trained models:

\begin{itemize}
    \item \textbf{Baseline (a):} The cosine similarity between quantized and unquantized activations fluctuates dramatically across layers, with multiple layers showing similarity scores approaching zero. This indicates that quantization errors accumulate and amplify through the network, causing the quantized model's internal representations to diverge completely from the full-precision model.

    \item \textbf{With Noise Injection (b):} The noise-trained model maintains consistently high cosine similarity ($>$0.95) across all layers, demonstrating that Jacobian-guided noise injection trains the model to be inherently robust to numerical perturbations introduced by quantization.
\end{itemize}

We also reproduced this in a PTQ setting with W4A4 Round-to-Nearest (RTN) quantisation with Siglip base 384 model. Figure 6 shows that errors accumulate over layers and eventually result in divergence of activation values. However, training with noise reduces this divergence and helps model consistently maintain high similarity with actual activation values.

\begin{figure}[h]
\centering
    \includegraphics[width=\linewidth]{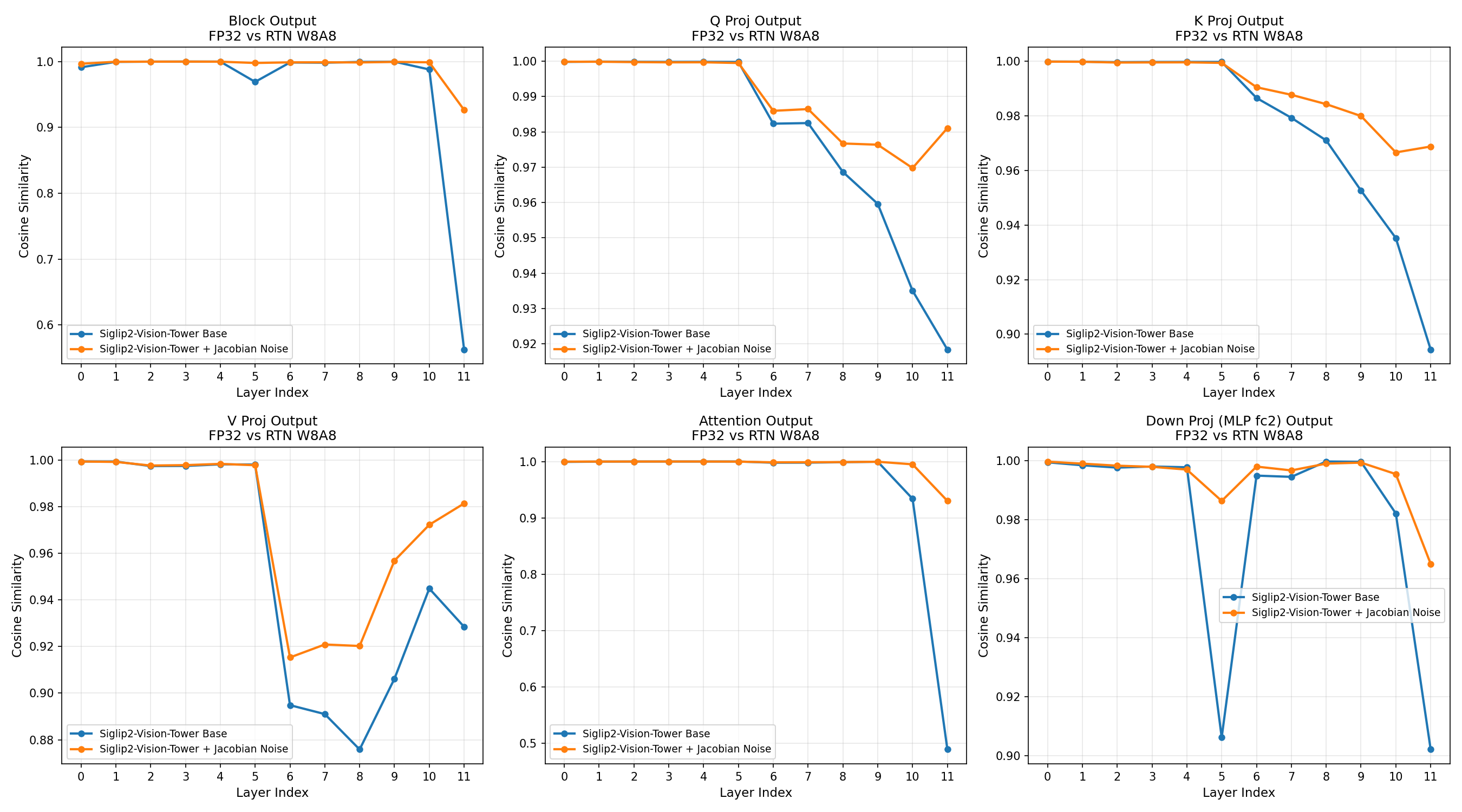}
    \label{fig:quant_cosine_similarity}
\caption{Attention maps for SigLIP (layer 10) Baseline vs baseline with noise. Noise-trained models exhibit more diffused attention patterns across key positions.}
\end{figure}

\textbf{Implications for Deployment.} The layer-wise similarity analysis provides direct evidence that softmax instability, as characterized in \cref{sec:methodology}, causes real deployment failures. When similarity drops to near-zero at intermediate layers, the error propagates and compounds through subsequent attention operations, leading to outputs that bear little resemblance to the full-precision model. Our noise injection method addresses this root cause by regularizing the attention mechanism during training, resulting in models that maintain consistent behavior when deployed on diverse hardware with varying numerical precision.

\end{document}